\documentclass{article}
\PassOptionsToPackage{numbers,sort&compress}{natbib}
\usepackage[final,main]{shared/neurips_2026}
\makeatletter
\renewcommand{\@notice}{}
\makeatother
\usepackage[T1]{fontenc}
\usepackage[utf8]{inputenc}
\usepackage{microtype}
\usepackage{booktabs}
\usepackage{array}
\usepackage{graphicx}
\usepackage{amsmath,amssymb}
\usepackage{enumitem}
\usepackage{siunitx}
\usepackage[hidelinks]{hyperref}
\usepackage{hyperxmp}
\hypersetup{
  hypertexnames=false,
  pdftitle={Refinement vs. Replanning: Inference Compute Allocation in Robot Policies},
  pdfauthor={Alba Maria Tellez Fernandez},
  pdfpublisher={MAAT Institute},
  pdfcontactemail={alba.tellez@maatinstitute.com},
  pdfcontacturl={https://amtellezfernandez.com/},
  pdfcopyright={Copyright 2026 Alba Maria Tellez Fernandez},
  pdflicenseurl={https://creativecommons.org/licenses/by/4.0/},
  pdfsubject={Post-Training Adaptation of Robot Foundation Models; NeurIPS 2026 workshop; deployment-time compute allocation for chunked robot policies},
  pdfkeywords={robot foundation models, vision-language-action models, VLA policies, post-training adaptation, deployment-time adaptation, inference compute, test-time compute, compute allocation, refinement vs. replanning, action chunking, replanning, closed-loop robot control, SmolVLA, LIBERO}
}
\usepackage{placeins}
\usepackage{float}
\title{Refinement vs. Replanning: Inference Compute Allocation in Robot Policies}
\author{\href{https://amtellezfernandez.com/}{Alba Mar\'ia T\'ellez Fern\'andez}\\
\href{https://maatinstitute.com/}{MAAT Institute}\\
\texttt{alba.tellez@maatinstitute.com}}
\begin{document}
\maketitle

\begin{abstract}
A frozen chunked robot policy can spend deployment-time inference compute in two coupled ways:
more work within each policy call to refine a predicted action chunk, or more
frequent policy calls to replan from fresh observations. We study this allocation
without updating model weights. Our evaluation procedure reports success with
both nominal compute rate and complete attempt cost, selects comparisons on
discovery tasks, and tests prespecified comparisons on held-out tasks. Comparisons
added after inspecting the held-out results only summarize observed patterns. We
demonstrate the procedure
with one frozen SmolVLA checkpoint in synchronous LIBERO simulation, using
denoising count as the refinement control and fixing chunk length at $K=50$.
Arithmetic and latency characterize cost differently: at $d=10$, 82.1\% of
counted arithmetic is independent of denoising, whereas a descriptive affine
fit to fixed-input host time assigns 17.7\% to its intercept. At $h=2$, $d=2$
succeeds on 139/160 attempts versus 122/160 for $d=5$ ($+10.6$ percentage
points; task-conditional 95\% interval: $+4.4$ to $+16.9$), while using 5.9\%
less arithmetic per control step. Executing $h=5$ rather than $h=2$ at $d=2$
yields 138/160 versus 139/160 successes and 60.7\% lower arithmetic per attempt;
equivalence is not established. The case study shows why deployment adaptation
must evaluate generation effort and feedback frequency jointly; it does not
define a universal setting.
\end{abstract}

\section{Introduction}

We can adapt a robot policy after training without changing its weights. At
deployment, a frozen chunked policy has two coupled compute decisions: how much
work each policy call spends generating an action chunk, and how often to call
the policy again from a fresh observation. We call the first decision
\emph{refinement} and the second \emph{replanning frequency}. Each policy call
predicts $K$ actions. The environment executes the first $h$, then the policy
receives a new observation and runs again. Thus, $h$ sets the policy-call
frequency: one call every $h$ executed actions. For SmolVLA, $d$ sets the number
of denoising steps within each call.

These choices couple computation and feedback. Additional refinement can alter
the predicted actions. A longer execution prefix amortizes a policy call, but it
delays feedback and exposes later actions from an older prediction to the
environment. Holding predicted chunk length fixed, we ask one allocation
question: \emph{under an explicit compute ceiling, how much computation should
each SmolVLA policy call spend on denoising refinement, and how often should the
policy run again from a fresh observation?} We select the combination that
maximizes closed-loop task success. Here, ``best'' means the highest
empirical success rate among configurations that satisfy the stated compute
ceiling. Counted arithmetic determines feasibility; we report nominal compute
rates and realized attempt costs separately. We also report measured host
latency, but it does not determine feasibility. We do not combine success and
cost into one score or measure memory use.

We study one frozen SmolVLA checkpoint \cite{smolvla} and fix predicted chunk
length at $K=50$. Our contribution is a controlled study of the allocation
between denoising work per policy call and policy-call frequency, together with
its effect on task success and complete attempt cost. We do not claim that the
selected allocation transfers to other checkpoints.

SmolVLA also provides an asynchronous inference mode that overlaps action
generation with action execution. We do not use that mode. In our synchronous
evaluation, the policy finishes generating all $K$ actions before the
environment executes the first $h$; the policy then receives a new observation
and replans. Fixing this order lets us compare refinement and replanning without
the additional effects of concurrent execution, observation staleness, or
buffering. The results provide a controlled baseline for a later asynchronous
study, but they do not identify the best asynchronous schedule or its
performance. We do not measure asynchronous throughput or physical-robot
latency.

We test 12 configurations in 480 discovery attempts, then seven settings in
1,120 held-out attempts on eight tasks excluded from selection. We chose the
refinement comparisons before the held-out evaluation. We added the
execution-prefix comparisons after inspecting those results, so they motivate
future tests rather than support a confirmed claim.

The contributions are:
\begin{itemize}[leftmargin=*,nosep]
    \item a compute-accounting formulation that connects SmolVLA's denoising
    work per policy call and its policy-call frequency at fixed chunk length;
    \item a controlled discovery and held-out evaluation that measures task
    success together with nominal compute rate and complete attempt cost; and
    \item evidence that most counted per-call arithmetic is independent of
    denoising, that fewer denoising steps improve held-out success in the
    primary comparison, and that success under less frequent replanning remains
    unresolved.
\end{itemize}

\section{Related Work}

\paragraph{Post-training allocation under a compute budget.}
Compute-optimal scaling studies ask how a fixed training budget should be divided
between model size and data \cite{chinchilla}; related embodied work studies
scaling in agent pretraining and world models \cite{pearcescaling}. Including
deployment demand can change the preferred allocation \cite{beyondchinchilla}.
These studies establish the need to state the budget and counting boundary. We
instead adapt the inference configuration of one frozen policy and evaluate
each allocation of deployment-time inference compute by closed-loop task success.

\paragraph{Computation used to generate an action.}
Diffusion Policy generates action sequences through iterative conditional
denoising \cite{diffusionpolicy}.
RoboMonkey spends additional inference compute on action candidates and
verification \cite{robomonkey}. This is one form of refinement: both proposal and
verification cost belong to the policy call. SmolVLA provides the flow-matching
policy used in our case study and also studies asynchronous inference
\cite{smolvla}. VLA-Perf examines how model
architecture, hardware, placement, networking, and asynchronous execution affect
VLA inference performance \cite{vlaperf}. These works show that arithmetic and
elapsed time depend on both the model and its execution system. Our case study
holds the checkpoint, hardware, and synchronous execution path fixed so that the
effects of refinement and execution prefix can be compared within one system.

\paragraph{Choosing when to replan.}
ACT predicts action chunks to reduce the effective decision horizon
\cite{act}; chunked action spaces have also been studied for reinforcement
learning \cite{qchunking}.
Adaptive Action Chunking uses action entropy to choose how much of a predicted
chunk to execute \cite{adaptivechunking}. DVAC uses variation observed during
denoising to adapt the execution prefix \cite{dvac}. ElegantVLA schedules
vision--language recomputation, reuse, and action generation during execution
\cite{elegantvla}. These methods change the schedule within an episode. We map
fixed refinement and prefix settings first, separate selection from held-out
evaluation, and report both nominal compute rate and realized attempt cost.
This design evaluates axes available to an adaptive method; it does not compare
a fixed deployment configuration with an online, within-episode schedule.

\section{Allocating Inference Compute Between Refinement and Replanning}
\label{sec:allocation}

We represent the post-training deployment configuration of a frozen chunked
policy by its model-specific refinement configuration $\mathbf r$, predicted
chunk length $K$, and execution prefix $h$. The following quantities connect
those controls to the two cost measures used in the evaluation.

At each policy query, $1\leq h\leq K$ actions are executed and the remaining
$K-h$ are discarded before replanning. For fixed or reference-calibrated call
cost, let $F_{\mathrm{gen}}(\mathbf r,K)$ be measured arithmetic under a stated
boundary. The nominal compute rate per simulated control step is
\[
 b(h,K,\mathbf r)=\frac{F_{\mathrm{gen}}(\mathbf r,K)}{h}.
\]
Let $p_{\mathrm{success}}(h,K,\mathbf r)$ denote expected closed-loop success under the
specified task distribution.
For a candidate set $\mathcal C$ and budget ceiling $B$, the allocation target is
\begin{equation}
 (h^\star,K^\star,\mathbf r^\star)(B)\in
 \arg\max_{(h,K,\mathbf r)\in\mathcal C:\,b(h,K,\mathbf r)\le B}
 p_{\mathrm{success}}(h,K,\mathbf r).
 \label{eq:allocation}
\end{equation}
A configuration need not exhaust the ceiling. Equation~\ref{eq:allocation}
defines a within-checkpoint deployment decision; paired rollouts estimate its
unknown success term.

\begin{figure}[H]
\centering
\includegraphics[width=\linewidth]{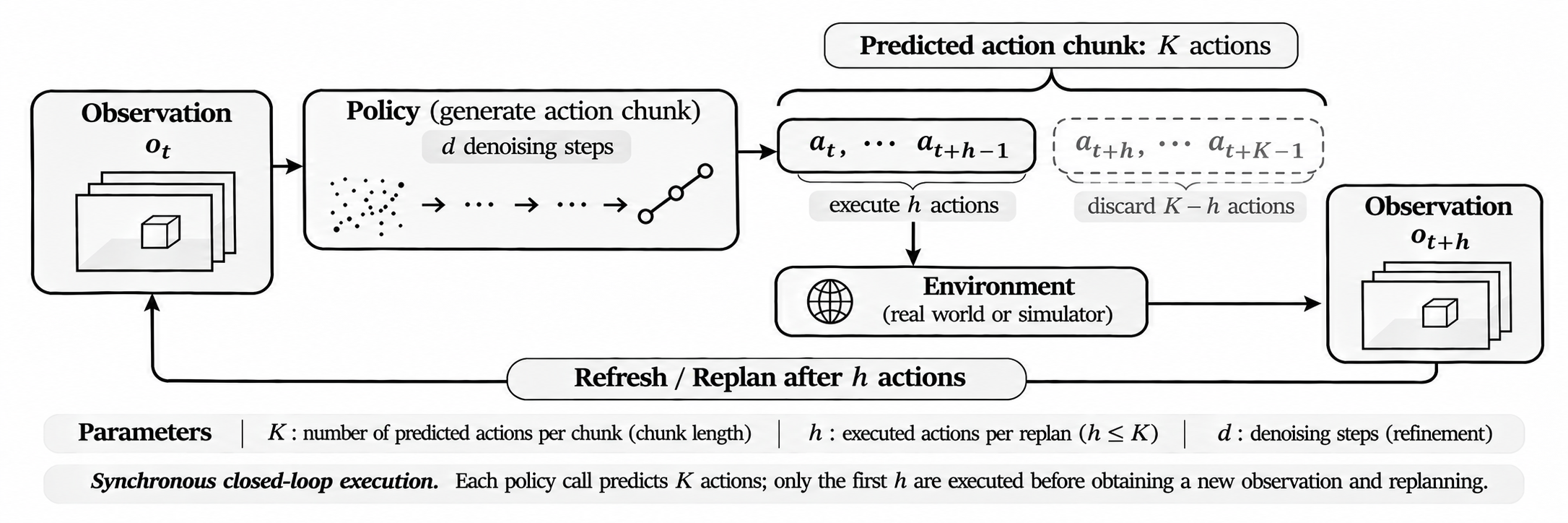}
\caption{Closed-loop execution in the SmolVLA case study. Each policy call
predicts $K$ actions after $d$ denoising steps, executes the first $h$, discards
the remainder, and replans from a new observation. Here $\mathbf r=(d)$ and
$K=50$.}
\label{fig:closed-loop-policy}
\end{figure}

In the SmolVLA instantiation, $K=50$ and $\mathbf r=(d)$, where $d$ is the
denoising count. Figure~\ref{fig:closed-loop-policy} shows this case-study execution loop.
For compactness below, $F_{\mathrm{gen}}(d)$ denotes
$F_{\mathrm{gen}}((d),50)$ and $b(h,d)$ denotes $b(h,50,(d))$.

We distinguish this rate from total arithmetic per attempt. If attempt $e$
executes $T_e$ actions, let $N_e=\lceil T_e/h\rceil$ be its number of policy
calls and let $F_{e,j}(\mathbf r,K)$ be the measured cost of call $j$. Attempt
$e$'s generated-chunk expenditure is
\begin{equation}
 C_e=\sum_{j=1}^{N_e}F_{e,j}(\mathbf r,K).
 \label{eq:episode-cost}
\end{equation}
The final generated chunk is charged even if only part of its execution prefix
is used. For the case study, each call has fixed cost for a given $d$, so this
reduces to $C_e=\lceil T_e/h\rceil F_{\mathrm{gen}}(d)$. The summed form also
covers variable-cost early exit, sampling, search, verification, or language
generation. All-attempt means include unsuccessful
episodes. Different trajectories and termination times can therefore change
expenditure at the same nominal rate. FLOPs, host elapsed time, and CUDA-stream
time retain separate units and measurement boundaries.

\section{SmolVLA Study Design and Configuration Selection}
\label{sec:method}

This section applies the allocation procedure to adapt the inference
configuration of one frozen checkpoint. Discovery maps the candidate settings
and selects the comparisons; the held-out evaluation then tests those
comparisons on tasks excluded from selection.

\subsection{Discovery and held-out evaluation}

We evaluate a public SmolVLA checkpoint on LIBERO \cite{libero}, with
\[
 h\in\{1,2,5\}, \qquad d\in\{2,5,10,20\}.
\]
The predicted chunk length is fixed at $K=50$, and the model's native deployment
uses $(h,d)=(1,10)$. We keep fixed the checkpoint, observations, preprocessing,
action normalization, control period, and task-specific episode horizons.

The discovery study evaluates these 12 settings on four Spatial tasks. Each
task has five initial states and
two rollout seeds, giving 40 paired task--state--seed identities. Every identity
receives all twelve settings in randomized order, for 480 attempts. Two discovery
tasks overlap pre-study pilot work; the other two were chosen before their study
outcomes were observed. This partial exposure is why discovery supports
selection rather than final evaluation.

After discovery, we freeze three denoising comparisons, one at each compute
ceiling. Their selected settings and comparators, plus native deployment, produce
seven distinct settings. The held-out evaluation applies all seven to eight Goal and
Object tasks excluded from selection. Each task has ten initial states and two
rollout seeds, giving 160 attempts per setting and 1,120 in total. Prior exposure
to these tasks during unrelated development is not assessed. No equivalence
margin was prespecified, so similar observed success cannot establish equivalent
quality. Appendix~\ref{app:analysis-status} distinguishes prespecified
comparisons from comparisons added after inspecting the results. The settings,
identities, stopping rules, inclusion criteria, and primary analysis were fixed
before held-out outcomes. After 114 attempts per primary setting, only the
execution order changed; the planned identities and analyses remained fixed.

Conditions are paired by initial state and rollout seed. They also share policy
noise while their queries occur at the same simulation step. Every denoising
setting traverses the same integration interval; changing $d$ changes only its
discretization. Consistency checks cover initial conditions, generated and executed
actions, and recorded outcomes. They do not provide an independent simulator
replay; Appendix~\ref{app:verification} states the validation boundary.

Success gives equal weight to each task; because every task has the same number
of states and seeds, this equals the overall success fraction. Descriptive 95\%
intervals use 20,000 bootstrap draws of states within the fixed tasks while
retaining paired seeds and settings. The intervals condition on the four
discovery tasks or eight held-out tasks, and secondary intervals are not
adjusted for multiple comparisons. Arithmetic includes every generated chunk,
including failed attempts and partly consumed final prefixes. Appendix~\ref{app:fixed-timing}
reports timing separately.

\subsection{A fixed-cost floor in the SmolVLA implementation}
\label{sec:fixed-cost-floor}

For the case-study checkpoint and reference input, generation cost separates
into a denoising-independent term and a term that grows linearly with $d$. To
rounding, the measured arithmetic is
\[
 F_{\mathrm{gen}}(d) \simeq 527.108 + 11.457\,d\quad\text{GFLOPs}
\]
at all six calibrated denoising settings. Native generation at $d=10$ costs
641.675 GFLOPs, of which 82.1\% is independent of $d$. Halving denoising from
ten to five saves 8.9\% of counted generation work; reducing it to two saves
14.3\%. Equivalently, relative to native deployment $(h,d)=(1,10)$,
\[
 \frac{b(h,d)}{b_{\mathrm{native}}}
 \simeq \frac{0.821+0.179(d/10)}{h}.
\]
Here $b_{\mathrm{native}}=b(1,10)$. Table~\ref{tab:confirmation-grid} reports
the measured version of this ratio in its ``Nominal rate ratio'' column for all
held-out settings; Appendix Table~\ref{tab:resources} reports the corresponding
discovery measurements. The displayed expression is a rounded summary of those
measured costs, not a separate estimate.

Profiling counts matrix products, convolutions, and attention across vision,
language prefill, and the action expert, using two FLOPs per multiply-add.
Embedding lookups, normalization, activations, elementwise arithmetic, and
external preprocessing are excluded. We therefore refer to \emph{dense neural
arithmetic}, rather than every scalar operation. Calibration uses one real
development observation and fixed tensor shapes; Appendix~\ref{app:generation-calibration}
gives the procedure.

We prespecify compute-rate ceilings of $0.25$, $0.50$, and $1.00$ times native.
Eligibility uses measured per-$d$ values, not the rounded affine expression.
For example, $(1,2)$ costs 0.857 times native rate and is ineligible at half-native
budget, whereas $(2,2)$ costs 0.429 times native. The denoising-independent work
alone excludes $h=1$ at half-native budget and $h\leq3$ at quarter-native budget.
A tighter budget can therefore force a longer execution prefix before any
quality-based choice is made. This arithmetic floor is specific to the
implementation and declared counting boundary; it is not a universal VLA law.

Calibration uses a fixed input shape. A separate audit using the language lengths
observed in the tasks leaves eligibility unchanged under the same absolute
ceilings (Appendix~\ref{app:shape-audit}). The reference-input arithmetic used
for selection remains distinct from task-specific arithmetic and elapsed time.
A descriptive affine fit to the six fixed-input host-time means gives
$T_{\mathrm{gen}}(d) \simeq 112.3 + 52.2\,d$ ms. Its intercept is 17.7\% of fitted
$d=10$ host time, compared with 82.1\% for counted arithmetic. The timing
intercept is an empirical fit under one host boundary, not a component
attribution. Consistently, reducing $d$ from 10 to 5 lowers measured host time by
40.5\% while counted arithmetic falls by 8.9\%
(Appendix~\ref{app:fixed-timing}); FLOPs are therefore not a latency proxy. The
cost model determines which settings satisfy a ceiling. The following experiments
test their closed-loop outcomes.

\subsection{Discovery: observed success is not monotonic in refinement}
\label{sec:discovery-quality}

Native $(1,10)$ succeeds in 25/40 discovery attempts. The highest observed count
is 28/40 for $(1,5)$; $(1,2)$ and $(2,2)$ each succeed in 26/40. At $h=5$,
success falls from 21/40 to 14/40 as $d$ increases from 2 to 20, with intermediate
counts of 18/40 and 16/40.
Appendix Figure~\ref{fig:frontier} shows every configuration.
These point estimates do not justify imposing monotonic improvement with
denoising. The small paired sample also does not certify equal quality for
configurations with similar observed counts.

Selection also depends on how cost is defined. Nominal arithmetic rate differs
from realized attempt expenditure, and arithmetic savings differ from elapsed
time. Replacing the arithmetic ceiling with a host-time ceiling changes which
prefixes are feasible. This sensitivity analysis does not alter the frozen
arithmetic-based experiment. Appendix~\ref{app:discovery-outcomes} gives the
full discovery results, and Appendix~\ref{app:timing-sensitivity} gives the
resource-coordinate analysis.

\subsection{Selection rules and prespecified comparisons}
\label{sec:prediction}

Discovery freezes $(5,2)$, $(2,2)$, and $(1,2)$ at the quarter-, half-, and
full-native ceilings, with $(5,5)$, $(2,5)$, and $(1,5)$ as their respective
comparators; $(2,2)$ versus $(2,5)$ is primary. Every selected setting is an
observed grid point, so the held-out evaluation tests transfer to new tasks rather than
extrapolation. Each pair varies denoising at a fixed prefix and compares two
feasible settings; the pair does not divide an identical compute rate.

At the primary ceiling, $(2,2)$ has the highest eligible discovery success.
Among the remaining settings, $(2,5)$ ties $(2,10)$ and wins the lower-cost
tie-break. A discovery-fitted logistic model and a simpler rule---fewest
denoising steps, then shortest feasible prefix---produce the same three
selections. Because the simpler rule was documented after data collection
began, the agreement is descriptive and does not show an advantage from fitting.
Appendix~\ref{app:selectors} gives the model, candidate set, and diagnostics.

\FloatBarrier
\raggedbottom
\section{Held-Out Success and Compute}
\label{sec:transfer}

\begin{figure}[H]
\centering
\includegraphics[width=\linewidth]{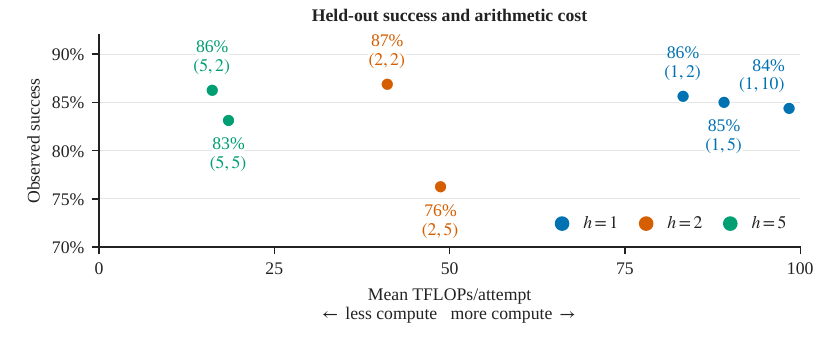}\par\vspace{-0.65em}
\caption{Held-out success against mean arithmetic per attempt for all seven
settings. Every setting has 160 attempts; labels give $(h,d)$ and observed
success. Color denotes the execution prefix $h$; every setting uses the same
circular marker. The success axis begins at 70\%.
Selection ceilings use nominal arithmetic per control step rather than the
attempt cost on this axis (Section~\ref{sec:fixed-cost-floor}); exact values
appear in Table~\ref{tab:confirmation-grid}.}
\label{fig:success-compute}
\end{figure}

The held-out evaluation applies seven settings to the same 160
task--state--seed identities, for 1,120 planned attempts. Three fixed-$h$
comparisons of $d=2$ against $d=5$ were prespecified, with $(2,2)$ versus
$(2,5)$ as primary. The replanning comparisons were added after the held-out
results were inspected and are interpreted as observed patterns. Figure~\ref{fig:success-compute}
shows the overall success--cost landscape; Table~\ref{tab:confirmation-grid}
gives the exact values. Appendix~\ref{app:verification} documents two removed
duplicate records without changing the planned totals.

\paragraph{Primary result: lower refinement improves success while reducing compute.}
The primary comparison is designed to isolate refinement: both settings predict
$K=50$ actions, execute two actions, and then replan from a new observation; only
the denoising count changes. Moving from $d=5$ to $d=2$ raises success from
122/160 to 139/160. Because both settings use the same task--state--seed
identities, the $10.6$-percentage-point increase is a within-identity difference,
not a comparison between different rollout samples. The paired outcomes contain
24 cases where only $d=2$ succeeds, seven where only $d=5$ succeeds, and 129
ties, giving the observed net gain of 17 successes. The task-conditional 95\%
interval, $[+4.4,\,+16.9]$, excludes zero for these eight tasks.

\begin{table}[H]
\centering\small
\caption{Held-out results for the seven SmolVLA settings. Each setting has 160
attempts across eight tasks. The predicted chunk length is fixed at $K=50$.
Nominal rate is counted arithmetic per control step relative to native $(1,10)$;
mean arithmetic per attempt charges every generated chunk.}
\label{tab:confirmation-grid}
\begin{tabular}{@{}ccrS[table-format=2.1]S[table-format=1.3]S[table-format=2.1]S[table-format=3.1]@{}}
\toprule
\multicolumn{2}{c}{Allocation} & \multicolumn{2}{c}{Success} & {Nominal} & \multicolumn{2}{c}{Mean per attempt}\\
\cmidrule(lr){1-2}\cmidrule(lr){3-4}\cmidrule(l){5-5}\cmidrule(l){6-7}
$h$ & $d$ & Count & {(\%)} & {rate ratio} & {TFLOPs} & {Steps}\\
\midrule
1 & 2 & 137/160 & 85.6 & 0.857 & 83.3 & 151.4\\
1 & 5 & 136/160 & 85.0 & 0.911 & 89.1 & 152.5\\
1 & 10 & 135/160 & 84.4 & 1.000 & 98.4 & 153.4\\
\addlinespace[3pt]
2 & 2 & 139/160 & 86.9 & 0.429 & 41.1 & 149.0\\
2 & 5 & 122/160 & 76.3 & 0.455 & 48.7 & 166.3\\
\addlinespace[3pt]
5 & 2 & 138/160 & 86.3 & 0.171 & 16.1 & 144.9\\
5 & 5 & 133/160 & 83.1 & 0.182 & 18.5 & 156.2\\
\bottomrule
\end{tabular}

\end{table}

\FloatBarrier

The compute reduction has two scales. Lowering $d$ reduces nominal arithmetic
per control step by 5.9\%. Mean arithmetic per complete attempt falls by more,
15.6\%, from 48.7 to 41.1 TFLOPs, because attempt expenditure also depends on
how many actions are executed and therefore how many policy calls are made. The
$d=5$ setting averages 166.3 executed steps, compared with 149.0 for $d=2$.
Thus, for this checkpoint at $h=2$, additional denoising is both less successful
and more expensive under the declared arithmetic boundary.

The result is consistent across the sampled tasks---seven favor $d=2$ and one
ties---but it is not a general denoising trend. At $h=1$ and $h=5$, the point
estimates also favor $d=2$, yet their intervals cross zero. The advantage may
therefore depend on how refinement interacts with replanning frequency. The
response model predicted the primary direction but understated its magnitude
($+4.0$ versus $+10.6$ points), so these data also do not demonstrate calibrated
forecasting.

\begin{figure}[H]
\centering
\includegraphics[width=\linewidth]{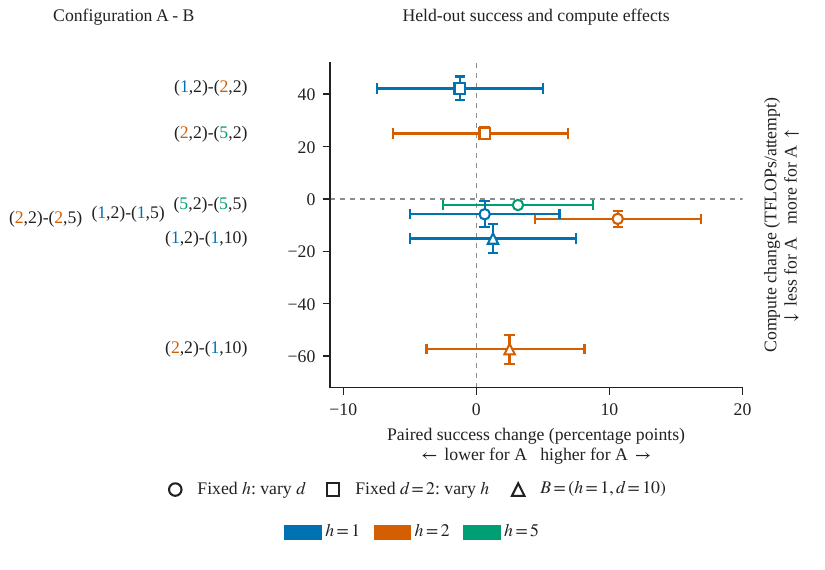}
\vspace{-8pt}
\caption{Joint success and compute effects for the seven held-out comparisons.
In each label, the first setting is A and the second is B. The marker shows how
A changes relative to B: moving right means that A succeeds more often, moving
left means that it succeeds less often, moving down means that A uses less
arithmetic per attempt, and moving up means that it uses more. The lower-right
quadrant therefore favors A on both measures. Horizontal arms are the
task-conditional 95\% intervals for success change; vertical arms are the paired
95\% intervals for compute change. Color denotes $h$ for configuration A,
matching Figure~\ref{fig:success-compute}; marker shapes identify the comparison
type, as shown below the plot. Exact definitions and counts appear in
Table~\ref{tab:confirmation-secondary}.}
\label{fig:confirmation-effects}
\end{figure}
\FloatBarrier
\paragraph{Replanning changes cost much more than observed success.}
At $d=2$, executing five rather than two actions per prediction reduces mean
arithmetic per attempt by 60.7\%, from 41.1 to 16.1 TFLOPs. Observed success
changes from 139/160 to 138/160, but the paired interval includes practically
meaningful effects in both directions and task-level directions are mixed.
Replanning after every action is also more expensive and shows no measured
success advantage over replanning after two actions. These comparisons establish
the cost differences under the declared counting boundary; they do not establish
equivalent quality or a general recommendation about replanning frequency.
Appendix~\ref{app:confirmation-details} gives their analysis status and the two
contrasts involving native $(1,10)$; Table~\ref{tab:confirmation-tasks} shows
task-level variation.

\paragraph{Evidence and scope.}
The prespecified evidence establishes the primary result for this checkpoint at
$h=2$. The replanning comparisons were selected after inspecting the results,
so their success differences require a future prespecified study. Neither varied
control isolates a single mechanism: changing denoising changes numerical
integration, while changing the prefix changes both observation age and executed
action positions. The study uses one checkpoint, twelve tasks, and synchronous
simulation; its intervals are conditional on the sampled tasks, and no
equivalence margin was prespecified. Counted arithmetic excludes some operation
classes and is not a latency proxy. Recorded outcomes pass source-tracing and
consistency checks, but the full held-out trajectories were not replayed in an
independent, version-matched simulator. Appendix~\ref{app:verification} defines
this validation boundary. Within these limits, the held-out evidence supports
lower refinement at $h=2$ and establishes the compute---but not quality---effects
of longer execution prefixes.

\FloatBarrier
\flushbottom
\section{Conclusion}

Adapting a frozen policy through its deployment configuration can change both
success and total compute. In the primary SmolVLA comparison, reducing
denoising from five to two steps improves held-out success while reducing
counted arithmetic by 5.9\% per control step and 15.6\% per attempt. The other
prespecified prefixes do not establish a uniform refinement effect. Executing
five rather than two actions before replanning produces the much larger compute
saving---60.7\% per attempt---but the available descriptive comparison does not
establish equivalent success. These results show why success, nominal rate, and
complete attempt cost must be evaluated together.

Why might fewer denoising steps also improve success? One plausible explanation
is that additional denoising does more than improve numerical precision: every
step applies a learned update, so approximation or calibration errors can
accumulate and move the first actions away from a more successful prediction.
At $h=2$, the policy executes only two of the $K=50$ predicted actions before
using a new observation, which may limit the value of further refining the full
chunk. This mechanism is consistent with the primary result, but the experiment
does not identify its cause. Testing it requires action-level comparisons across
denoising counts and trajectory analysis around the states where outcomes
diverge.

The contribution is a controlled way to study this allocation for a frozen
SmolVLA checkpoint: vary denoising work and policy-call frequency, measure every
call under an explicit boundary, choose comparisons before held-out evaluation,
and report success with realized attempt cost. The selected setting is specific
to this checkpoint and counting boundary.

The present study keeps $K=50$ throughout, leaving predicted chunk length as a
central untested allocation lever; with the tested prefixes, 90--98\% of every
prediction is discarded before the next query. The next study should
prespecify a quality-preservation test of the cheaper $(5,2)$ setting, vary $K$,
and separate observation freshness from action-position effects. Repeating the
same within-model design across checkpoints and before and after weight-based
post-training adaptation would determine whether those adaptations also change
the preferred deployment configuration. This is the evidence needed to decide
when compute should refine the current prediction and when it should produce a
new prediction from fresher feedback.

\section*{Data and Code Availability}
Code, compact outcomes, and runnable examples can be found in the
\href{https://github.com/maatinstitute/ReVa}{\texttt{maatinstitute/ReVa}}
repository. The examples reproduce the reported study and illustrate how to
adapt the same evaluation to other models and checkpoints. Full simulator
rollouts, model weights, and private runtime artifacts are not redistributed.

\bibliographystyle{unsrtnat}
{\raggedright\bibliography{references}}
\clearpage
\appendix

\section{Discovery and Resource Measurements}
\label{app:calibration}

Appendices A--C follow the study chronology. This section reports discovery
outcomes, generation-cost calibration, fixed-input timing, sensitivity to the
resource coordinate, and the language-shape audit. Appendix~\ref{app:selectors}
documents frozen selection; Appendix~\ref{app:held-out-details} reports and
verifies held-out outcomes.

\subsection{Discovery outcomes}
\label{app:discovery-outcomes}

{\raggedright
The checkpoint is \texttt{HuggingFaceVLA/smolvla\_libero}, revision
\texttt{6721902bc4d61e50a3bfdb11dfb4cb626f05d102}. Predicted chunk length is 50;
native execution uses $h=1,d=10$. Discovery uses Spatial tasks 5--8, initial
states 10--14, and rollout seeds 2000 and 2001. The 20 task--state blocks
contain both seeds and all 12 configurations. Discovery bootstrap resampling uses
20,000 draws with analysis seed 20260913.\par}

\begin{table}[htbp]
\centering\small
\caption{Discovery successes on Spatial tasks 5--8. Each task contributes five
states and two seeds per configuration. The nominal rate ratio is
$F_{\mathrm{gen}}(d)/(hF_{\mathrm{gen}}(10))$, not realized attempt cost.
The best observed count is 28/40 at $(1,5)$; no monotonic denoising trend is
assumed.}
\label{tab:discovery-grid}
\begin{tabular}{@{}ccrrrrrS[table-format=1.3]@{}}
\toprule
\multicolumn{2}{c}{Allocation} & \multicolumn{4}{c}{Spatial task successes (each /10)} & & {Nominal}\\
\cmidrule(lr){1-2}\cmidrule(lr){3-6}
$h$ & $d$ & Task 5 & Task 6 & Task 7 & Task 8 & Total & {rate ratio}\\
\midrule
1 & 2 & 5 & 7 & 4 & 10 & 26/40 & 0.857\\
1 & 5 & 5 & 8 & 6 & 9 & 28/40 & 0.911\\
1 & 10 & 5 & 7 & 6 & 7 & 25/40 & 1.000\\
1 & 20 & 5 & 6 & 6 & 8 & 25/40 & 1.179\\
\addlinespace[3pt]
2 & 2 & 5 & 7 & 4 & 10 & 26/40 & 0.429\\
2 & 5 & 4 & 6 & 4 & 9 & 23/40 & 0.455\\
2 & 10 & 4 & 6 & 4 & 9 & 23/40 & 0.500\\
2 & 20 & 4 & 7 & 4 & 9 & 24/40 & 0.589\\
\addlinespace[3pt]
5 & 2 & 6 & 5 & 4 & 6 & 21/40 & 0.171\\
5 & 5 & 2 & 5 & 5 & 6 & 18/40 & 0.182\\
5 & 10 & 2 & 4 & 5 & 5 & 16/40 & 0.200\\
5 & 20 & 2 & 3 & 4 & 5 & 14/40 & 0.236\\
\bottomrule
\end{tabular}

\end{table}

\begin{figure}[htbp]
\centering
\includegraphics[width=\linewidth]{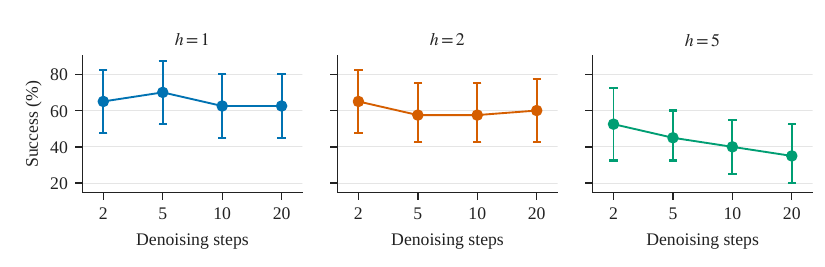}
\caption{Discovery success by executed prefix $h$, with 40 attempts per
configuration on four Spatial tasks. Connected estimates make the within-prefix
response to denoising visible; whiskers are descriptive, task-conditional 95\%
paired state-bootstrap intervals. The $d$ axes are categorical.}
\label{fig:frontier}
\end{figure}

\FloatBarrier
\subsection{Generation calibration and discovery resources}
\label{app:generation-calibration}

Calibration uses one real observation from Spatial task 0, initial state 0,
and common explicit noise. For each $d\in\{2,3,5,8,10,20\}$, profiling records
solver timesteps and dense operation counts, separately from elapsed timing.
Each setting receives 20 warm-up generations, followed by three randomized
timing blocks of 20 measured calls per setting. The rollout evaluator also
performs 20 warm-up generations once per configuration.

Host generation timing brackets prediction and completion synchronization;
CUDA events measure the current-stream prediction interval. The outer rollout
clock additionally includes observation processing, action handling,
simulation/rendering, and output handling. Model loading, environment creation,
reset, and warm-up are measured separately. Process sampling occurs approximately
every five seconds; the largest recorded gap during discovery is below
5.03 seconds. Display processes remain on the measurement GPU.

\begin{table}[htbp]
\centering\small
\caption{Discovery resource measurements. Generated-chunk totals use the declared
dense-arithmetic boundary and charge partially executed final prefixes.
Episode means include unsuccessful attempts. $(2,2)$ has 26/40 observed
successes at 42.9\% of the native nominal arithmetic rate.}
\label{tab:resources}
\begin{tabular}{@{}ccS[table-format=3.1]S[table-format=3.1]S[table-format=3.1]r@{}}
\toprule
\multicolumn{2}{c}{Allocation} & {Nominal / step} & \multicolumn{2}{c}{Mean per attempt} & Success\\
\cmidrule(lr){1-2}\cmidrule(lr){3-3}\cmidrule(lr){4-5}
$h$ & $d$ & {GFLOPs} & {TFLOPs} & {Rollout (s)} & Count\\
\midrule
1 & 2 & 550.0 & 90.6 & 39.5 & 26/40\\
1 & 5 & 584.4 & 91.7 & 61.8 & 28/40\\
1 & 10 & 641.7 & 109.6 & 111.9 & 25/40\\
1 & 20 & 756.2 & 129.1 & 199.8 & 25/40\\
\addlinespace[3pt]
2 & 2 & 275.0 & 45.0 & 21.8 & 26/40\\
2 & 5 & 292.2 & 53.4 & 38.5 & 23/40\\
2 & 10 & 320.8 & 57.8 & 61.6 & 23/40\\
2 & 20 & 378.1 & 67.8 & 107.8 & 24/40\\
\addlinespace[3pt]
5 & 2 & 110.0 & 20.5 & 12.7 & 21/40\\
5 & 5 & 116.9 & 23.8 & 20.2 & 18/40\\
5 & 10 & 128.3 & 27.2 & 32.0 & 16/40\\
5 & 20 & 151.2 & 33.1 & 55.6 & 14/40\\
\bottomrule
\end{tabular}

\end{table}

Calibration requires approximately 291 seconds and discovery 31,245 seconds
(8.68 hours) of host elapsed time, including setup and bookkeeping. This search
expenditure is additional to subsequent deployments. The experiment does not
estimate when deployment amortizes this search cost.

All measurements use an RTX 4060 Ti. Process samples taken approximately every
five seconds show no unrelated sampled CUDA compute during calibration and
discovery; display processes remain. Coverage is incomplete during the held-out
evaluation, so its aggregate rollout times are descriptive. Host and CUDA-stream
generation intervals are measured separately from the outer synchronous rollout
interval, and warm-up is excluded.

\subsection{Fixed-input generation timing}
\label{app:fixed-timing}

\begin{table}[htbp]
\centering\small
\caption{Fixed-input full-generation calibration. Each setting has 60 calls in
three randomized blocks. Range gives the minimum and maximum host block means,
not a confidence interval. Host time falls more steeply with $d$ than counted
arithmetic. Host and CUDA columns average separate measurements of all 60 calls; the extra decimals distinguish their means, not timing accuracy.}
\begin{tabular}{@{}cS[table-format=3.1]S[table-format=4.3]S[table-format=4.3]r@{}}
\toprule
$d$ & {GFLOPs} & {Host (ms)} & {CUDA (ms)} & Host block range (ms)\\
\midrule
2 & 550.0 & 217.891 & 217.862 & 214.5--220.1\\
3 & 561.5 & 269.990 & 269.970 & 265.3--272.6\\
5 & 584.4 & 374.321 & 374.300 & 366.8--378.1\\
8 & 618.8 & 529.746 & 529.726 & 519.9--536.1\\
10 & 641.7 & 629.624 & 629.604 & 623.2--642.2\\
20 & 756.2 & 1158.635 & 1158.608 & 1133.5--1172.6\\
\bottomrule
\end{tabular}
\end{table}

The calibration stores full-generation intervals and global operator arithmetic
counts, not separate vision, language-prefill, or action-expert timings and
arithmetic totals. These records cannot attribute elapsed sensitivity to a
particular component, memory traffic, omitted operation classes, synchronization,
or software overhead. CUDA-event intervals are elapsed stream measurements,
not summed active-kernel durations. No component-level decomposition is inferred.

\subsection{Resource-coordinate sensitivity}
\label{app:timing-sensitivity}

Arithmetic savings and elapsed savings differ. Mean rollout time is 111.9 seconds
per attempt for native deployment, 61.8 seconds for $(1,5)$, and 21.8 seconds for
$(2,2)$. Their mean generated-chunk expenditures are 109.6, 91.7, and 45.0 TFLOPs,
respectively. These are observed all-attempt means under the same task allocation;
they include different trajectories and outcomes. All discovery configurations
appear in Table~\ref{tab:resources}.

There is also an ordering reversal: $(1,5)$ uses more arithmetic per attempt
than $(2,20)$ (91.7 versus 67.8 TFLOPs), but less rollout time (61.8 versus
107.8 seconds). These means include different trajectories. On a fixed input,
reducing $d$ from 10 to 5 lowers mean generation host time from 629.6 to
374.3 ms (40.5\%), despite only 8.9\% less counted arithmetic. The calibration
therefore separates arithmetic savings from elapsed savings without trajectory
differences, but it does not resolve their component-level cause. FLOP and
latency constraints can therefore select different settings.

As a descriptive sensitivity analysis, define calibrated host-time rate
$b_T(h,d)=\overline{T}_{\mathrm{gen}}(d)/h$, the amortized fixed-input generation
time per simulated control step, and normalize it by native host-time rate.
For $(1,2)$, $(2,2)$, and $(5,2)$, the normalized arithmetic rates are
0.857, 0.429, and 0.171, while the normalized host-time rates are 0.346, 0.173,
and 0.069. Consequently, $(1,2)$ is ineligible at a half-native arithmetic
ceiling but eligible at a separately normalized half-native host-time ceiling.
These resource constraints admit different replanning schedules; the choices do
not establish quality optima. The frozen experiment remains arithmetic-based.

\begin{table}[htbp]
\centering\small
\caption{Smallest execution prefix feasible for $d=2$ under separately
normalized arithmetic and fixed-input host-time ceilings. The resource
coordinate changes feasibility; these choices do not compare outcome quality.}
\label{tab:resource-coordinate-choices}
\begin{tabular}{@{}lcc@{}}
\toprule
Ceiling relative to native & Arithmetic constraint & Host-time constraint\\
\midrule
0.25 & $(5,2)$ & $(2,2)$\\
0.50 & $(2,2)$ & $(1,2)$\\
1.00 & $(1,2)$ & $(1,2)$\\
\bottomrule
\end{tabular}

\end{table}

\subsection{Language-shape transport audit}
\label{app:shape-audit}

The profiled batch contains one sample, two $3\times360\times360$ image tensors,
an eight-dimensional state tensor, and language-token and attention-mask tensors
of length 20. The checkpoint exposes seven-dimensional actions and generates a
50-action chunk. Evaluation fixes batch size, image preprocessing, state/action
dimensions, and predicted chunk length. The tokenizer pads language only to the
longest sequence in a batch. Reconstructing the declared task strings with the pinned
tokenizer gives 15--17 tokens for discovery and 5--12 for held-out evaluation, versus 20
in calibration. We profile every distinct evaluated length for
$d\in\{2,5,10\}$, preserving all other shapes, and test eligibility against
the prespecified absolute ceilings. Table~\ref{tab:shape-audit} reports the range
over evaluated lengths 5--17 against the original length-20 value.

\begin{table}[htbp]
\centering\small
\caption{Dense generation arithmetic across evaluated language shapes. Budget
tests retain the prespecified absolute ceilings; none of the reported held-out
configurations changes eligibility.}
\label{tab:shape-audit}
\begin{tabular}{@{}crS[table-format=3.1]c@{}}
\toprule
$d$ & \shortstack{Evaluated lengths\\GFLOPs (range)} & {\shortstack{Length 20\\GFLOPs}} & \shortstack{Eligibility\\changes}\\
\midrule
2  & 539.7--547.9 & 550.0 & 0\\
5  & 573.5--582.2 & 584.4 & 0\\
10 & 629.8--639.3 & 641.7 & 0\\
\bottomrule
\end{tabular}
\end{table}

In particular, $(2,10)$ remains within the original half-native ceiling for all
evaluated lengths; shorter inputs reduce rather than increase its counted work.
The audit changes neither policy execution nor the frozen selections.

\FloatBarrier
\section{Selection and Diagnostic Analyses}
\label{app:selectors}

\subsection{Frozen selection procedure}

We compare two logistic response-model families. A compute-only model uses a
quadratic in log normalized rate. An allocation model uses $\log h$,
$\log(d/10)$, both squared terms, and their interaction. Each family is fitted
with ridge penalty 1 or 10. Leave-one-task-out log loss selects the model, which
is then refitted on the discovery set. No individual seed repeat is held out
while another repeat from the same task remains in training.

The candidate set is $h\in\{1,2,3,5\}$ crossed with $d\in\{2,3,5,8,10,20\}$:
the 12 discovery configurations and 12 untested configurations. Their calibrated costs determine
eligibility; the fitted model selects the highest predicted success under each
ceiling. Ties use compute and lexicographic order.

\begin{table}[htbp]
\centering\small
\caption{Discovery-based selector choices.
The learned choices and comparisons were fixed before held-out evaluation; additional
selector baselines were documented after collection began.
The empirical selector uses the measured grid; other selectors use the declared
candidate set. The simple rule exactly matches the learned choices. The selected
rate ratio is the learned recommendation's nominal arithmetic rate
divided by the native rate. Prespecified comparators
and predicted gaps appear in Table~\ref{tab:confirmation-predictions};
the half-native row is primary.
}
\label{tab:selectors}
\begin{tabular}{@{}lccccS[table-format=1.3]@{}}
\toprule
 & \multicolumn{4}{c}{Selector choice $(h,d)$} & {Learned selector}\\
\cmidrule(lr){2-5}
Ceiling ratio & Learned & Simple & Empirical & Compute-only & {Rate ratio}\\
\midrule
0.25 & $(5,2)$ & $(5,2)$ & $(5,2)$ & $(5,20)$ & 0.171\\
0.50\ (primary) & $(2,2)$ & $(2,2)$ & $(2,2)$ & $(2,10)$ & 0.429\\
1.00 & $(1,2)$ & $(1,2)$ & $(1,5)$ & $(1,10)$ & 0.857\\
\bottomrule
\end{tabular}

\end{table}

\begin{table}[htbp]
\centering\small
\caption{All prespecified denoising comparisons: selected minus comparator,
holding $h$ fixed within each row. Predicted and observed gaps and intervals are
percentage points. The half-native row is primary; secondary intervals are
descriptive, task-conditional, and unadjusted.}
\label{tab:confirmation-predictions}
\begin{tabular}{@{}lccS[table-format=+2.1]S[table-format=+2.1]l@{}}
\toprule
 & \multicolumn{2}{c}{Setting $(h,d)$} & \multicolumn{3}{c}{Success difference (pp)}\\
\cmidrule(lr){2-3}\cmidrule(l){4-6}
Ceiling ratio & Selected & Comparator & {Predicted} & {Observed} & {95\% interval}\\
\midrule
0.25 & $(5,2)$ & $(5,5)$ & +6.5 & +3.1 & $[-2.5,\,+8.8]$\\
0.50\ (primary) & $(2,2)$ & $(2,5)$ & +4.0 & +10.6 & $[+4.4,\,+16.9]$\\
1.00 & $(1,2)$ & $(1,5)$ & +2.2 & +0.6 & $[-5.0,\,+6.3]$\\
\bottomrule
\end{tabular}

\end{table}

\subsection{Selector diagnostics}

The simple rule defined in Section~\ref{sec:prediction} reproduces every learned
recommendation (Table~\ref{tab:selectors});
held-out outcomes cannot distinguish these selectors at the tested budgets.
The compute-only choices use the original ridge-10 fit. Its quarter- and
half-budget choices are absent from the held-out plan, so this evaluation
cannot compare those decisions on held-out tasks.

The selected allocation model has log loss 0.7048, compared with 0.7067 for the
best compute-only model. It does not beat the trivial constant-$p=0.5$ baseline,
which scores 0.6931 under the same mean natural-log binary cross-entropy; an
intercept-only model fitted separately on
each training fold scores 0.7170. The response-model probabilities should
therefore not be read as calibrated held-out-task forecasts. The primary test
evaluates a recommendation shared with simpler selectors, so it does not
establish an additional benefit from model-based ranking.

\FloatBarrier
\section{Held-Out Outcomes and Verification}
\label{app:held-out-details}

\subsection{Complete held-out outcomes}
\label{app:confirmation-details}

This subsection first reports success for every held-out task and setting, then
reports the four comparisons added after collection began. Of those four, two
isolate the execution prefix, one isolates denoising, and one changes both.
The held-out evaluation uses Goal and Object tasks 5--8, states 20--29, and
rollout seeds 3000 and 3001. Each of its 80 task--state blocks contains both
seeds. All seven configurations appear in every block.

\begin{table}[htbp]
\centering\small
\caption{Held-out successes by task and configuration. Each task row has
20 attempts: ten paired initial states, each with two rollout seeds. The total
aggregates 160 attempts per configuration. The primary
$(2,2)$ setting exceeds $(2,5)$ on seven tasks and ties on one.}
\label{tab:primary-tasks}
\label{tab:confirmation-tasks}
\begin{tabular}{@{}lS[table-format=3.0]S[table-format=3.0]S[table-format=3.0]S[table-format=3.0]S[table-format=3.0]S[table-format=3.0]S[table-format=3.0]@{}}
\toprule
 & \multicolumn{3}{c}{$h=1$} & \multicolumn{2}{c}{$h=2$ (primary pair)} & \multicolumn{2}{c}{$h=5$}\\
\cmidrule(lr){2-4}\cmidrule(lr){5-6}\cmidrule(l){7-8}
Task & {$d=2$} & {$d=5$} & {$d=10$} & {$d=2$} & {$d=5$} & {$d=2$} & {$d=5$}\\
\midrule
Goal 5 & 14 & 15 & 14 & 16 & 14 & 18 & 16\\
Goal 6 & 18 & 17 & 17 & 18 & 15 & 19 & 18\\
Goal 7 & 20 & 20 & 20 & 19 & 18 & 18 & 18\\
Goal 8 & 19 & 18 & 18 & 18 & 14 & 18 & 15\\
\addlinespace[3pt]
Object 5 & 20 & 19 & 19 & 20 & 20 & 20 & 20\\
Object 6 & 19 & 18 & 19 & 20 & 17 & 18 & 18\\
Object 7 & 9 & 9 & 10 & 10 & 7 & 7 & 8\\
Object 8 & 18 & 20 & 18 & 18 & 17 & 20 & 20\\
\midrule
Total (/160) & 137 & 136 & 135 & 139 & 122 & 138 & 133\\
\bottomrule
\end{tabular}

\end{table}

\begin{table}[htbp]
\centering\small
\caption{Post-outcome descriptive comparisons using all 160 attempts per
setting. These contrasts were defined after collection began and were not
prespecified or fully outcome-blind. Intervals are task-conditional,
unadjusted, and do not test whether two settings preserve quality within a
prespecified margin.}
\label{tab:confirmation-secondary}
\begin{tabular}{@{}ccS[table-format=+2.1]l@{}}
\toprule
Setting A & Setting B & {A$-$B (pp)} & 95\% interval (pp)\\
\midrule
$(2,2)$ & $(1,10)$ & +2.5 & $[-3.8,\,+8.1]$\\
$(2,2)$ & $(5,2)$ & +0.6 & $[-6.3,\,+6.9]$\\
$(1,2)$ & $(2,2)$ & -1.3 & $[-7.5,\,+5.0]$\\
$(1,2)$ & $(1,10)$ & +1.3 & $[-5.0,\,+7.5]$\\
\bottomrule
\end{tabular}

\end{table}
\FloatBarrier

The four post-outcome comparisons use
20,000 paired state-bootstrap resamples and analysis seed 20260913.
Table~\ref{tab:confirmation-secondary} reports their success differences and
uncertainty intervals, while
Table~\ref{tab:confirmation-grid} gives their all-attempt resource means.
In particular, the $(2,2)$ minus $(5,2)$ expenditure difference is
$+25.0$ TFLOPs per attempt, with descriptive interval $[+22.8,\,+27.5]$.
This interval describes generated-chunk arithmetic, including the different
trajectory lengths; it is not an elapsed-time comparison or evidence of
equivalent quality.

\FloatBarrier
\subsection{Primary contrast robustness, expenditure, and runtime}
\label{app:primary-costs}

The primary selected configuration $(2,2)$ has a higher success count than
$(2,5)$ on seven tasks and ties on one (Table~\ref{tab:primary-tasks}).
There are 24 paired wins and seven
losses. In a descriptive leave-one-task-out sensitivity analysis, the
equally weighted difference over the remaining seven tasks stays positive,
ranging from 9.3 to 12.1 percentage points. This post-outcome analysis does not
extend inference to a new task population. The ReVa example provides the paired episode-level identities needed to
reproduce the primary counts and comparison.

\begin{table}[htbp]
\centering\small
\caption{Descriptive primary expenditure differences, selected minus comparator
at $h=2$, including all attempts. Intervals use the paired state bootstrap
conditional on the eight tasks and were added after outcomes were observed.
Timing intervals summarize sampling variation in the observed data; incomplete
process-monitoring coverage remains a limitation. Differences use unrounded
means and may differ from subtraction of the displayed values.}
\label{tab:primary-costs}
\begin{tabular}{@{}lS[table-format=3.1]S[table-format=3.1]S[table-format=-2.1]l@{}}
\toprule
Mean per attempt & {$d=2$} & {$d=5$} & {Difference} & 95\% interval\\
\midrule
Reference TFLOPs & 41.1 & 48.7 & -7.6 & $[-10.6,\,-4.5]$\\
Observed rollout (s) & 19.0 & 34.3 & -15.2 & $[-17.2,\,-13.3]$\\
Control steps & 149.0 & 166.3 & -17.3 & $[-27.9,\,-6.5]$\\
\bottomrule
\end{tabular}

\end{table}

\paragraph{Runtime and environment.}
\begin{table}[H]
\centering\small
\caption{Runtime and evaluation conditions for the held-out evaluation.}
\label{tab:runtime}
\begin{tabular}{@{}>{\raggedright\arraybackslash}p{0.24\linewidth}>{\raggedright\arraybackslash}p{0.71\linewidth}@{}}
\toprule
Setting & Value\\
\midrule
Hardware & RTX 4060 Ti, 16 GB; driver 595.84\\
Runtime & Python~3.12.3; PyTorch~2.7.1+cu126; CUDA~12.6;\newline NumPy~2.2.6; MuJoCo~3.8.1\\
\addlinespace[3pt]
Evaluation & Batch 1; synchronous simulation; nominal 20 Hz (50 ms control period)\\
Episode horizons & Spatial and Object: 280 control steps; Goal: 300\\
\addlinespace[3pt]
Precision controls & Autocast and TF32 disabled; deterministic algorithms enabled\\
Attention and reuse & Eager float32 attention; within-generation KV reuse\\
Compilation/warm-up & Model compilation disabled; 20 warm-up generations per condition\\
\addlinespace[3pt]
Evaluator & Pinned evaluator revision\\
\bottomrule
\end{tabular}
\end{table}

The available measurements do not characterize numerical precision and
attention execution for every model component. Table~\ref{tab:runtime} supports
the stated model configuration, but not component- or kernel-level claims.

{\small
\subsection{Measurement verification}
\label{app:verification}

The evaluation data reconcile planned identities, paired states, executed
actions, and outcomes. The checks
in Table~\ref{tab:verification-summary} constrain recording and execution
errors; they do not constitute an
independent outcome replay in a separate simulator.

\begin{table}[H]
\centering\small
\caption{Measurement-verification checklist.}
\label{tab:verification-summary}
\begin{tabular}{@{}p{0.25\linewidth}p{0.67\linewidth}@{}}
\toprule
Check & Evidence\\
\midrule
Canonical coverage & 1,120 held-out attempts and 171,805 actions; the
primary subset remains 320 attempts and 50,451 actions.\\
Pairing and dispatch & Planned identities, paired initial states and
observations, generation counts, denoising counts, and applied actions reconcile.\\
Excluded attempts & Two repeated completed attempts are checked and excluded;
one interrupted ten-action record has no completed outcome and is excluded.\\
Outcome path & All 1,122 completed attempts and 172,243 steps reconcile;
generated actions match simulator actions and recorded outcome fields agree.\\
Shared predicate & Reward and success both use LIBERO's task-goal evaluation,
traced to BDDL predicates, so their agreement is a consistency check.\\
Native parity & One Spatial episode matches the upstream path in initial state,
observation, 70-action sequence, length, and outcome.\\
Remaining boundary & Independent, version-matched replay and direct terminal-state
inspection remain outside this audit.\\
\bottomrule
\end{tabular}
\end{table}
\FloatBarrier
\subsection{Analysis status}
\label{app:analysis-status}

The three refinement comparisons, paired identities, stopping rules, inclusion
criteria, and primary analysis were fixed before held-out outcomes. After
114 attempts per primary setting, the remaining execution order changed, while
the planned identities and analyses stayed fixed. The refinement analyses use
seed 20260912. Contrasts and diagnostics defined after collection began use seed
20260913 and are descriptive. The response model was not refitted on held-out
outcomes.

\paragraph{Reproducibility materials.}
The \href{https://github.com/maatinstitute/ReVa}{\texttt{maatinstitute/ReVa}}
repository contains the reproducibility materials and runnable examples. They
show how to reconstruct the reported aggregate results and how to extend the
evaluation to other models, checkpoints, and simulator settings. The original
study's full action records, model weights, images, and runtime artifacts are
not redistributed.
}

\end{document}